\documentclass[letterpaper, 10 pt, conference]{ieeeconf}  

\IEEEoverridecommandlockouts                              
\usepackage{amsmath} 
\usepackage[pdftex]{graphicx}
\usepackage[dvipsnames]{xcolor}
\usepackage{graphicx}

\title{\LARGE \bf
CDR Based Trajectories: Tentative for Filtering Ping-pong Handover
}

\author{Joonas L\~{o}mps$^{1}$, Artjom Lind$^{1}$ and Amnir Hadachi$^{1}$
\thanks{$^{1}$ J. L\~{o}mps, A. Lind and A. Hadachi are with ITS Lab within Institute of Computer Science at the Faculty of Science and Technology, University of Tartu, 51014 Tartu, Estonia.
        {\tt\small joonas.lomps@ut.ee}}%
}

\begin{document}

\maketitle
\thispagestyle{empty}
\pagestyle{empty}

\begin{abstract}

Call Detail Records (CDRs) coupled with the coverage area locations provide the operator with an incredible amount of information on its customers' whereabouts and movement. 
Due to the non-static and overlapping nature of the antenna coverage area there commonly exist situations where cellphones geographically close to each other can be connected to different antennas due to handover rule - the operator hands over a certain cellphone to another antenna to spread the load between antennas. Hence, this aspect introduces a ping-pong handover phenomena in the trajectories extracted from the CDR data which can be misleading in understanding the mobility pattern. To reconstruct accurate trajectories it is a must to reduce the number of those handovers appearing in the dataset. This letter presents a novel approach for filtering ping-pong handovers from CDR based trajectories. Primarily, the approach is based on anchors model utilizing different features and parameters extracted from the coverage areas and reconstructed trajectories mined from the CDR data. Using this methodology we can significantly reduce the ping-pong handover noise in the trajectories, which gives a more accurate reconstruction of the customers' movement pattern.

\end{abstract}



\section{INTRODUCTION}

In the last year, we witnessed an increase in the scientific research contribution concerning the usage of mobile data or call detail records (CDRs) in many applications related to mobility aspect and intelligent transportation systems such as trajectory reconstruction \cite{c1, c2}, mobility hubs discovery\cite{c3}, traffic monitoring \cite{c5}, mobility episodes detection\cite{c4}, travel time estimation \cite{c6}, etc. From this perspective, it is clear that mobile data has a great potential in depicting mobility patterns \cite{c8} from macroscopic perceptive and to some extent from microscopic level \cite{c7}. However, during the process of manipulating the CDR data and extracting trajectories from it, many researchers noticed that the handover phenomena, which is the process of transferring a connection from one cell to another without disrupting the session, create anomalies and misleading information about the movement of the mobile users which affect the quality of the mobility patterns \cite{c9}. In particular the frequent handovers between two cells also known as "ping-pong handover" occurs to be the most troublesome. The phenomena are well known and are caused by fluctuations in signal strength of two cells covering the same area with subscriber located on the boundaries of both cells. \\
Numerous approaches have been proposed to reduce the “ping-pong handover” distortions appearing in the trajectories extracted from CDR data. In \cite{c10} the authors presented a time-dependent weight approach where they concluded that it was very difficult to have a reliable trajectory-based CDR data. Some studies used a more straightforward method by eliminating or disregarding all the events where the device has registered in a ping-pong manner (ABA is transformed to AB) \cite{c11} other researchers proposed a set of rules for different recurrent sequences and how to modify them to reflect a more reasonable traverse pattern of a mobile device during trips \cite{c12}.\\
Another technique has been investigated in \cite{c13}, the approach has two major steps the first is detecting the ping-pong handover and the second one is the replacement process. The detection is executed based on accumulating and spotting the repeated sequence of different cells in the trajectory throughout time. The replacement is performed based on three different approaches. The first one is a representative technique which focuses on replacing the whole ping-pong sequence by the cell symbol that has the highest number of visits recorded. However, the approach does not consider if the cells are neighbouring cells or not (coverage overlap between the cells). The second method is a limits technique that replaces the whole ping-pong sequence by the entry cell and the exits cell which solves the weakness of the previous methods to some extent. The last technique is a hybrid one where the two first approaches are mixed. The results demonstrated that the hybrid technique was the closest to reality. \\
Filtering ping-pong handover can be very challenging due to the topology and signal propagation within the network coverage since it has this tendency to fluctuate and change due to many factors such as weather and obstacles. In some case, as it was proposed in \cite{c14} the filtering process is done by focusing on the trajectories extracted from the CDR data and applying a split process when the sequences are not unique. Then, the middle cell is included in both sequences. This approach is also adopted by \cite{c15} with slight difference since it considers the time change in the occurrence of the events. \\
In \cite{c17}, the authors presented a novel approach for detecting handover based on trajectory data mining. The idea is to build upon spotting the frequent patterns from a large historical dataset of moving objects. The algorithm performance was evaluated based on simulated data and demonstrated that the proposed methods were capable of reducing the ping-pong handover, especially, within the microcells in the mobile networks. A different approach was proposed in \cite{c16}, to handle the ping-pong phenomena by care-of-address used in the previous access network when the device is connected. Then, to detect it is very simple is by discovering duplicated address and process validation of the new care-of-address to be sure. However, this method is developed based on additional information not only CDR data.\\
Hence, in this paper, we are proposing a new approach to filter ping-pong handover phenomena from trajectories extracted from CDR data. The proposed model is based on anchors that are defined in such a manner to consider a variety of features concerning CDR data and mobility patterns to discover and filter the ping-pong anomalies from the trajectories.

\section{PROBLEM STATEMENT}
Cellular networks or mobile networks are distributed over land areas through cells where each cell includes a fixed location transceiver known as a base station. Together, these cells provide radio coverage over larger geographical areas. The cellular networks are usually portrayed in a honeycomb pattern, which might be the case for larger rural areas where the network usage is limited due to the number of users, whereas for urban areas it commonly happens that the same geographical area is covered by multiple cells, all different sizes and capacity. 

To enable users equipment to communicate even if the equipment itself is moving through cells during transmission, the networks use a crucial functionality called handover. This is a process in telecommunication and mobile communication in which an ongoing call or a data session is transferred to a new cell to keep the signal uninterrupted when a mobile user moves between different cells. 
\\
Unfortunately, moving from one cell to another is not the only reason the network can initiate a handover from one cell to another, it can also be caused by fluctuations of the cell coverage area, interference, weather, the load on the cell, users predicted usage of the service, tall buildings and other reasons. These reasons are most commonly the reason for ping-pong handover where the user is switched back or forth between some cells, causing irregularities in the data. We have also noticed several handovers to cells that according to the cell coverage plan should not be close to the area, but are on the receiving end of a handover, we call these cases hops. These hops are caused by cells over radiating and the fluctuations of the cell coverage area due to weather conditions and they insert fake movement into the trajectory of the user and need to be removed for correct trajectory estimation.
\\
We propose a novel anchors model utilizing different features and parameters extracted from the coverage area and reconstructed trajectories mined from CDR data to filter out ping-pong handovers and hops from the sequential data.

\section{PROPOSED METHODOLOGY}

As described in the previous section the methodology aims to filter out the cases of ping-pong handovers, as well as hops to cells that are for some reason radiating further than expected. To this end, we have devised these eight anchors that decide whether the destination event should be accepted or not.\\
The first event will be accepted by default, unless it is not present in the cell coverage plan, as we have nothing to compare it with after this we follow this logic, which is also described in figure \ref{algo}:

\begin{figure}
\includegraphics[width=0.5\textwidth]{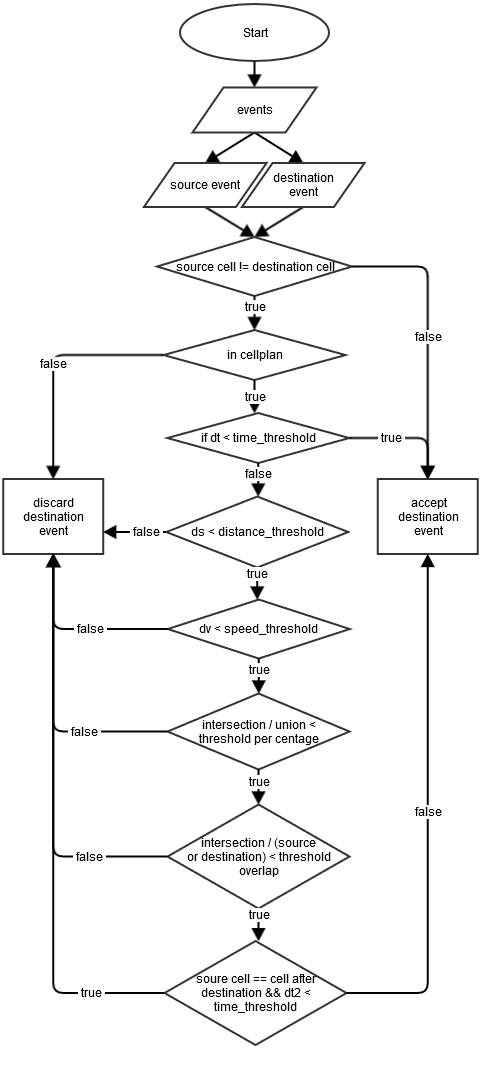}
\caption{Overview of the algorithm to filter events}
\centering
\label{algo}
\end{figure}

\begin{enumerate}
  \item Check that the source event is not from the same cell that the destination event. If they are not the same move on to the next anchor, otherwise accept the destination cell.
  \item We check if the destination cell is present in the cell coverage plan. If the destination cell is not present in the coverage plan we discard the event and compare the following event with the same source event. Reason being that the cell coverage plan is in constant change and unfortunately we do not have the resources and/or access to keep it up to date.
  \item We compare the times of the two events and if the $\Delta t < time\_threshold$ then we move on to the next anchor, otherwise accept the event and move to the next event. For our data threshold of 5-10 minutes worked the best. The reasoning behind this is that in those 5-10 minutes person using a vehicle in a city can move up to 3-6 kilometres or on the freeway up to 6-11 kilometres, which would make the entropy of the new location too large to be compared with the anchor method.
  \item Next we make sure that the distance from the source cell to the destination cell divided the distance from the source cells centroid to the destination cell centroid is below the $distance\_threshold$ (20\% in our case) we move to the next anchor, otherwise, we disregard the destination event. This anchor removes most of the hops too far away cells that are caught due to the fluctuations of the cells coverage area. Additionally, normal handovers from one cell to a neighbouring cell due to users movement do not get affected by this anchor.
  \item Followed by calculating the average speed needed to move from the source cell centroid to the destination cell centroid disregarding the topology in the given time between the two events. If the average speed $\Delta v < speed\_threshold$ we move to the next anchor, otherwise we disregard the destination event, as it should not be possible to move that fast, making it highly probable that it is a case of a hop or a ping-pong handover. We have chosen 90km/h as the $speed\_threshold$.
  \item Next the proportion of the coverage area of the two cells intersection to their union is compared to a threshold percentage. If it is below the set threshold ($similarity\_threshold$ = 50\% in our case), we move on to the next anchor, otherwise discard the event. Doing this allows us to discover handover events where both of the cells cover a very similar area, making the handover needed on the networks side of the view, but not the trajectory wise, most commonly it happens where the user spends a lot of time, e.g. home or work.
  \item This is followed by the comparison of the area of the intersection of the two cells with source and destination cell areas. If either of them is percentually larger than a given threshold ($covered\_threshold$ = 80\% in our case) we disregard the destination event, otherwise, we move to the last anchor point. This removes the cases where there are handovers from a smaller cell to a larger cell that almost entirely covers the latter or vice versa.
  \item Finally, we check for the cell pattern "ABA", meaning if the event after the current destination event is from the same cell as the source event and that the time difference of the last two events is below the  $time\_threshold$ in anchor 3. If that is the case, we classify it as ping-pong and discard that event, otherwise, we accept the destination event.
\end{enumerate}

These anchors let us filter out most of the ping-pong and hop handovers from the CDR data. 

\section{EXPERIMENTS, RESULTS AND DISCUSSION}
In order to test our developed methodology, we collected our own dataset using a mobile application to periodically log the GPS locations of the phone as well as all the location update events it receives. The GPS coordinates will be used as ground truth and the location update events will serve as the CDR data. \\
The phone was using a navigation application while travelling via car. It was also connected to all the usual social media applications commonly used by a large demographic of smartphone users e.g. Facebook, Messenger, Instagram. This setup will simulate a real-world use case and provide us with a close to real data since those applications periodically pull updates from the internet, which all generate a point of data at the operators' side. \\
Our collection was split into two different trajectories in order to get a better understanding of the problematic locations on the highway: first started in Tartu city, used the main highway to go to Tallinn city and then some movement inside the town itself, another one was the way back from Tallinn to Tartu city. In total the two trajectories form a dataset spanning over 55 hours, has 1757 GPS locations and 1174 location update events from 252 unique cells.\\

\begin{table}[!thb]
\centering
\begin{tabular}{c|r|r|r|r}
Trajectory              & \multicolumn{2}{c|}{1}                                        & \multicolumn{2}{c}{2}                                       \\ \hline
Dataset                 & \multicolumn{1}{c|}{Original} & \multicolumn{1}{c|}{Filtered} & \multicolumn{1}{c|}{Original} & \multicolumn{1}{c}{Filtered} \\ \hline
Location updates        & 898                           & 296                           & 276                           & 76                           \\ \hline
Unique cells            & 203                           & 131                           & 90                            & 54                           \\ \hline
Highest cell frequency & 69                            & 16                            & 15                            & 5                           
\end{tabular}
\caption{Comparisons of the statistics of the original and filtered trajectories}
\label{statistics}
\end{table}

Using the proposed methodology on the collected datasets resulted in filtering 802 events which removed a total of 80 unique cells from the datasets. The most frequent cells blinked 69 and 62 times whereas for the filtered case they were only seen 16 and 15 times respectively. It is also worth to mention that these cells cover the areas where the user spent the most time with the application running. Statistics are also displayed in table \ref{statistics} and table \ref{reasons} reflects the number of events each anchor accepted or discarded.

\begin{figure}[!thb]
\includegraphics[width=0.5\textwidth]{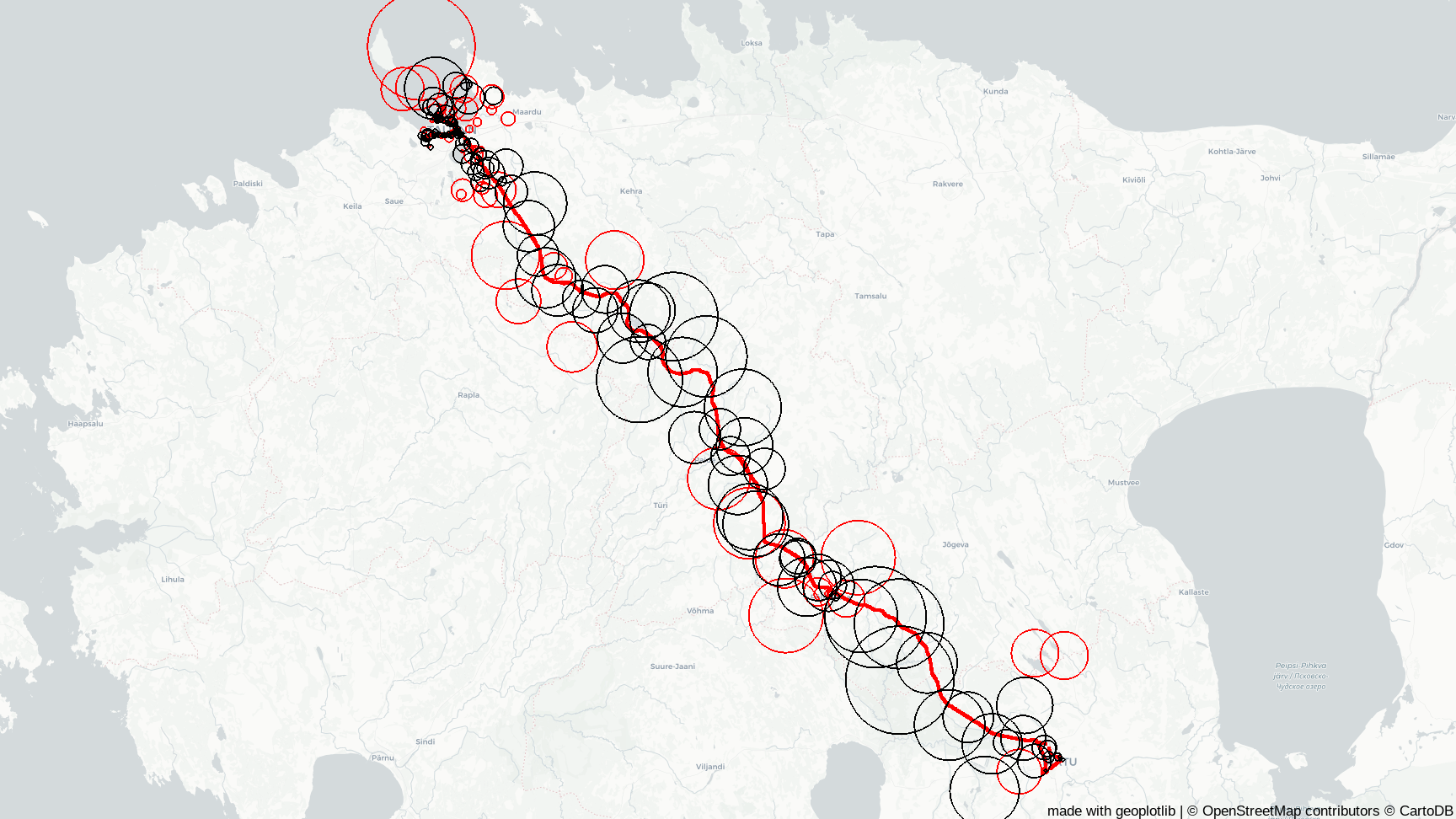}
\caption{Filtering effect of the methodology. Red dots indicate the recorded GPS locations and the circles are the associated CDR based trajectory - red circles indicate the cells that disappeared from the trajectory due to filtering the events and black circles are the ones remaining after filtering.}
\centering
\label{traj}
\end{figure}

\begin{figure}[!thb]
\includegraphics[width=0.5\textwidth]{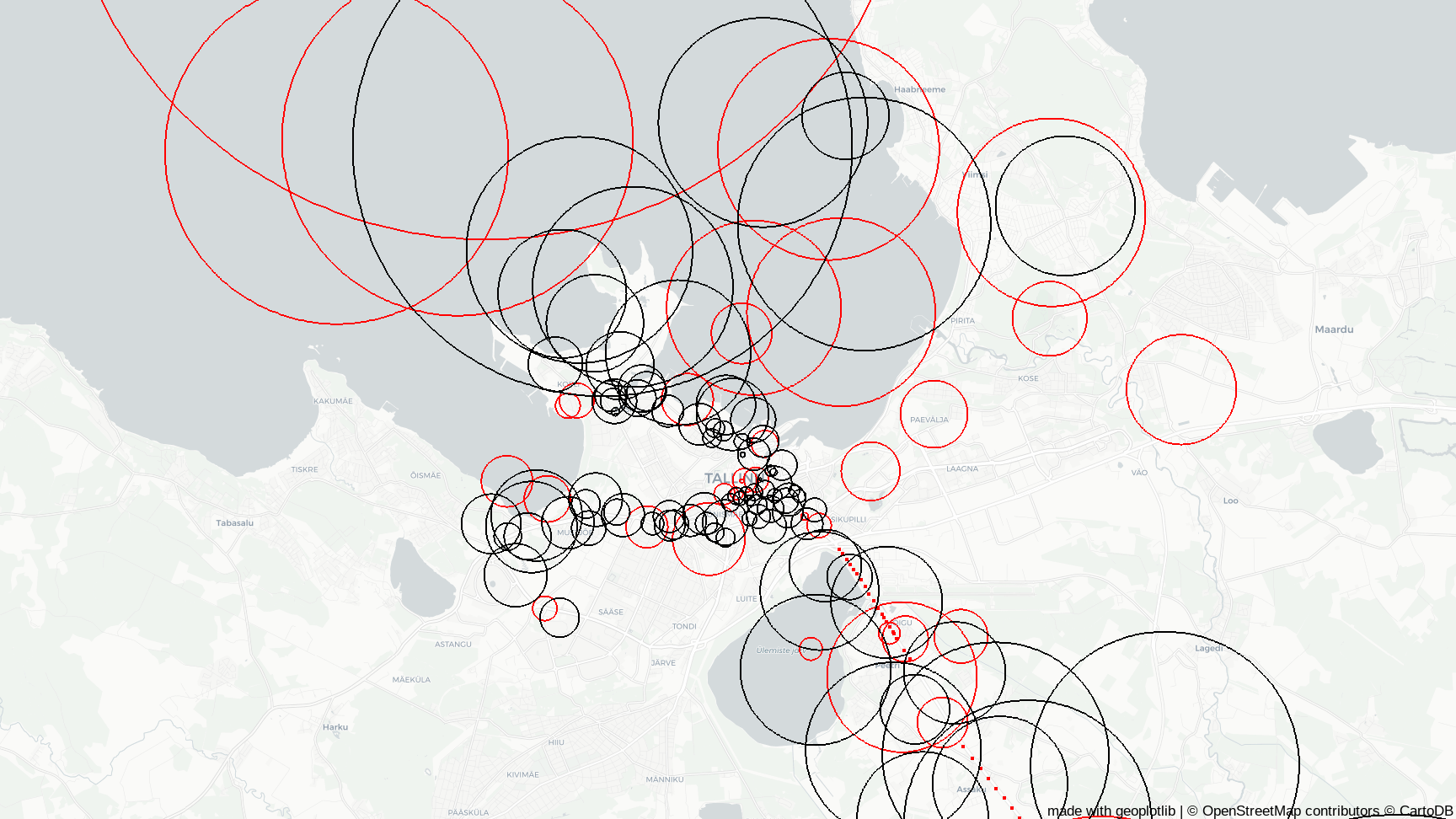}
\caption{Filtering effect of the methodology in an urban area (Tallinn). Red dots indicate the recorded GPS locations and the circles are the cells associated to the CDR based trajectory - red circles indicate the cells that disappeared from the trajectory due to filtering the events and black circles are the ones remaining after filtering.}
\centering
\label{tallinn}
\end{figure}

\begin{table}[!thb]
\centering
\begin{tabular}{r|r|r|r|r}
\multicolumn{1}{l|}{}       & \multicolumn{2}{c|}{Trajectory 1}                                            & \multicolumn{2}{c}{Trajectory 2}                                           \\ \hline
\multicolumn{1}{c|}{Anchor} & \multicolumn{1}{c|}{Accepted} & \multicolumn{1}{c|}{Discarded} & \multicolumn{1}{c|}{Accepted} & \multicolumn{1}{c}{Discarded} \\ \hline
1                           & -                                    & 48                                    & -                                    & 1                                    \\ \hline
2                           & -                                    & 69                                    & -                                    & 9                                    \\ \hline
3                           & 10                                   & -                                     & 1                                    & -                                    \\ \hline
4                           & -                                    & 230                                   & -                                    & 27                                   \\ \hline
5                           & -                                    & 246                                   & -                                    & 140                                  \\ \hline
6                           & -                                    & 19                                    & -                                    & 13                                   \\ \hline
7                           & -                                    & 38                                    & -                                    & 11                                   \\ \hline
8                           & -                                    & 0                                     & -                                    & 0                                   
\end{tabular}
\caption{Number of events accepted or discarded by certain anchors with the following threshold values $time\_threshold$ = 10 minutes (3),  $distance\_threshold$ = 20\% (4), $speed\_threshold$ = 90km/h (5), $similarity\_threshold$ = 50\% (6) and $covered\_threshold$ = 80\% (7)}
\label{reasons}
\end{table}
Figures \ref{traj} and \ref{tallinn} show the effects of our anchors based filtering algorithm in trajectory wide view and a more detailed view of an urban city center scenario. 
It can be seen that on the freeway several cells that do not contain the GPS points are removed. Same can be noticed for the urban area. Additionally, there are two clusters on the highway that contain several cells that were excluded from the trajectory although covering the GPS points. \\
For evaluation, we created two ground truth sets using the GPS and location update events collected for the trajectories using the temporal plane to associate a GPS location to every event. Then, for each event, the distance from the cell centroid to the associated GPS location was calculated and compared to the cell radius. For the first ground truth, we used a strict rule that the GPS location has to be in the range of the cell radius. For the second one, we took into account the fact that the cells coverage area tend to fluctuate due to environmental conditions and allowed the distance from the centroid to be $1.2 * r$, where $r$ is the cell radius. 

\begin{table}[!thb]
\centering
\begin{tabular}{c|r|r|r|r}
                    & \multicolumn{2}{c|}{Trajectory 1} & \multicolumn{2}{c}{Trajectory 2} \\ \hline
Groudtruth (GT)      & \multicolumn{1}{c|}{1} & \multicolumn{1}{c|}{2} & \multicolumn{1}{c|}{1} & \multicolumn{1}{c}{2} \\ \hline
Unique GT events     & 380             & 421             & 185             & 205             \\ \hline
Unique Filter events & 296             & 296             & 76              & 76              \\ \hline
Unique GT cells     & 115             & 128             & 54              & 58              \\ \hline
Unique Filter cells & 131             & 131             & 54              & 54              \\ \hline
Matching cells      & 95              & 106             & 41              & 43              \\ \hline
Not in GT cells     & 36              & 25              & 16              & 11              \\ \hline
Not in Filter cells & 20              & 22              & 16              & 15              \\ \hline
Precision           & 0.725           & 0.809           & 0.759           & 0.796           \\ \hline
Recall              & 0.826           & 0.828           & 0.759           & 0.741          
\end{tabular}
\caption{Evaluation paramaters and results}
\label{eval}
\end{table}

\begin{figure}[!thb]
\includegraphics[width=0.5\textwidth]{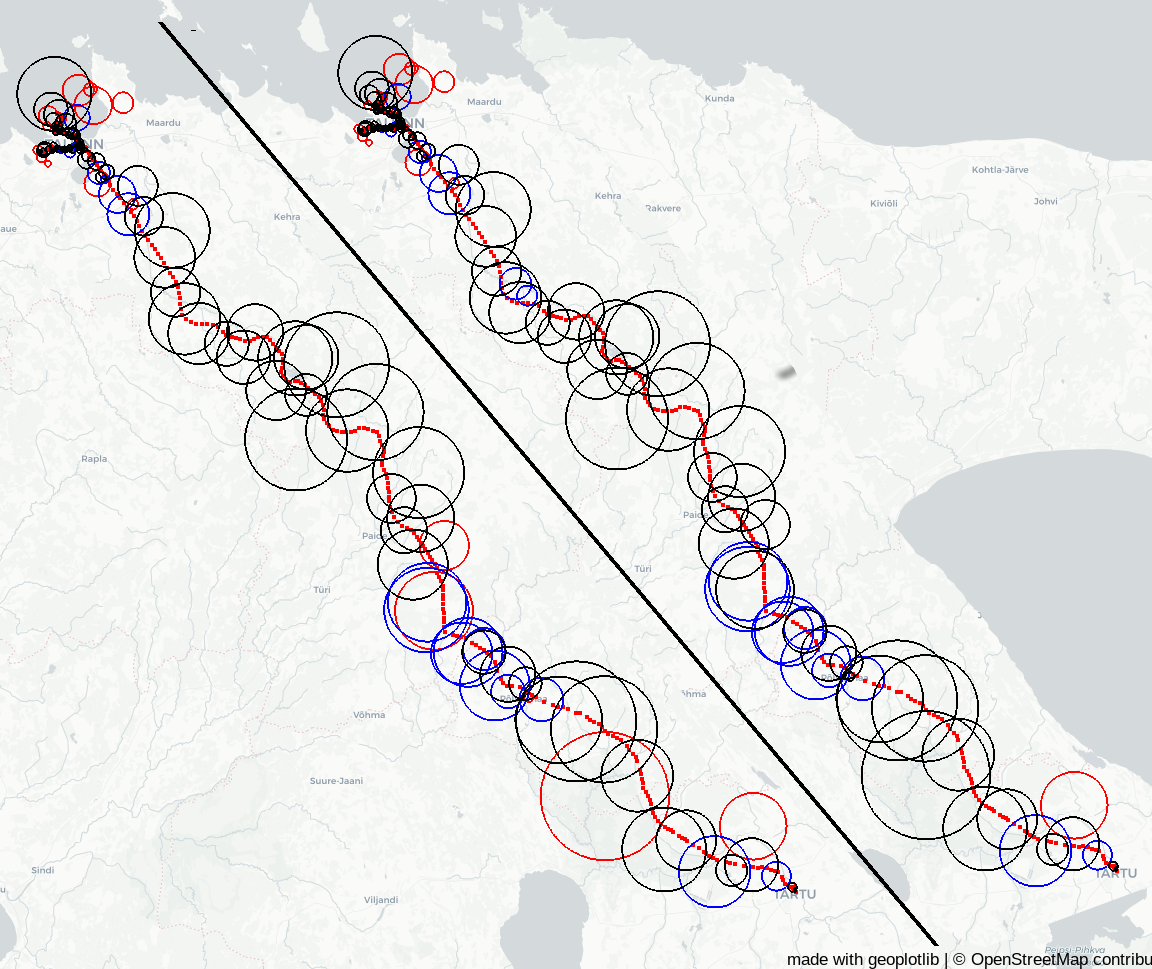}
\caption{Trajectory 1, ground truth 1 (left) vs ground truth 2 (right). Red dots indicate the recorded GPS locations and the circles are the cells associated to the CDR based trajectory - black circles indicate the cells that are present in both sets, the ground truth and filtered set, red circles indicate the cells that present only in the filtered set and blue ones only the ones in the ground truth.}
\centering
\label{gt1gt2}
\end{figure}

\begin{figure}[!thb]
\includegraphics[width=0.5\textwidth]{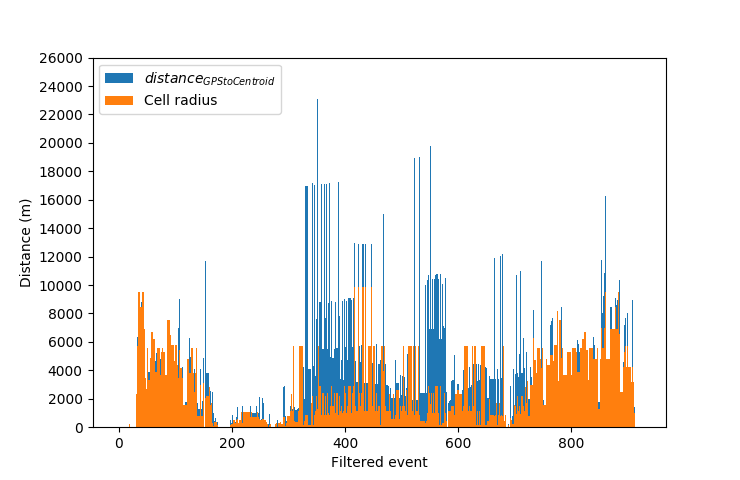}
\caption{Comparison of the distances from events cell centroid to its associated ground truth GPS location (blue) and the cells' radius (orange). Missing distances indicate cells for which there was no coverage information.}
\centering
\label{comparison}
\end{figure}

Table \ref{eval} reflects the results of the evaluation and event-based comparison can be seen on figure \ref{comparison}. Latter clearly shows cases where the GPS location distance from the cell centroid is a multiple of the cell radius. The evaluation shows that the precision of the anchor-based filtering increased when the cell coverage area fluctuation is taken into account. The overall difference between cells included by ground truth 1 and 2 can be seen in figure \ref{gt1gt2}. \\
There are still cases where handover to a far away from cell happens that are not filtered but with further parameter tuning and investigation into those specific cases might give us a hint of how to remove those. Also, there exist some false positive cases where events of which its cell that covers the area of the GPS location are removed, but as the graph does not show the temporal aspect it could be an anomaly or not. Additionally, looking back at a previously discarded event when processing the current event might lead to the removal of false-positive cases.

\section{CONCLUSION}
The preliminary study focuses on presenting a new concept of filtering ping-pong handover events from CDR trajectory data. 
The proposed method uses anchors method with eight points, taking into account the time, intersection, speed, coverage, similarity, and sequence. Using these, it filters out events it decides that are either ping-pong handover or hops. The results show that our concept allows us to discover and remove a great proportion of events that are not intersecting with the ground truth path. We also propose additions to the methodology to increase the accuracy of the discarding of hop events and reduce the number of false-positive removals. 


\addtolength{\textheight}{-12cm}   




\section*{ACKNOWLEDGMENT}
The authors gratefully acknowledge the contribution of Tele2 Eesti for their help in providing the data through the project "Population Movement Analytics, Monitoring and Prediction Algorithms". This project and research is supported by Archimedes Foundation and Mooncascade O\"U under the Framework of Support for Applied Research in Smart Specialization Growth Areas.


\bibliographystyle{IEEEtran}
\bibliography{ref}

\end{document}